\documentclass[authoryear,final,3p,times,twocolumn]{elsarticle}
\usepackage{graphicx}
\usepackage{amssymb}
\usepackage{lineno}
\usepackage[figuresright]{rotating}
\usepackage{multirow}
\usepackage[table]{xcolor}
\usepackage{arydshln}
\usepackage{amsmath}
\usepackage[colorlinks]{hyperref}

\journal{ISPRS Journal of Photogrammetry and Remote Sensing}
\begin{document}
\begin{frontmatter}


\title{Learning from Multimodal and Multitemporal Earth Observation Data for Building Damage Mapping}



\author[label1]{Bruno Adriano}
\author[label1,label2]{Naoto Yokoya}
\author[label1]{Junshi Xia}
\author[label3]{Hiroyuki Miura}
\author[label4]{Wen Liu}
\author[label5]{Masashi Matsuoka}
\author[label6]{Shunichi Koshimura}

\address[label1]{Geoinformatics Unit, RIKEN Center for Advance Intelligence Project, Japan}
\address[label2]{Complexity Science and Engineering, Graduate School of Frontier Sciences, the University of Tokyo, Japan}
\address[label3]{School of Advanced Science and Engineering, Hiroshima University, Japan}
\address[label4]{Graduate School of Engineering, Chiba University, Japan}
\address[label5]{Department of Architecture and Building Engineering, Tokyo Institute of Technology, Japan}
\address[label6]{International Research Institute of Disaster Science, Tohoku University, Japan}

\begin{abstract}
Earth observation (EO) technologies, such as optical imaging and synthetic aperture radar (SAR), provide excellent means to monitor ever-growing urban environments continuously. Notably, in the case of large-scale disasters (e.g., tsunamis and earthquakes), in which a response is highly time-critical, images from both data modalities can complement each other to accurately convey the full damage condition in the disaster's aftermath. However, due to several factors, such as weather and satellite coverage, it is often uncertain which data modality will be the first available for rapid disaster response efforts. Hence, novel methodologies that can utilize all accessible EO datasets are essential for disaster management. In this study, we have developed a global multisensor and multitemporal dataset for building damage mapping. We included building damage characteristics from three disaster types, namely, earthquakes, tsunamis, and typhoons, and considered three building damage categories. The global dataset contains high-resolution (HR) optical imagery and high-to-moderate-resolution multiband SAR data acquired before and after each disaster. Using this comprehensive dataset, we analyzed five data modality scenarios for damage mapping: single-mode (optical and SAR datasets), cross-modal (pre-disaster optical and post-disaster SAR datasets), and mode fusion scenarios. We defined a damage mapping framework for the semantic segmentation of damaged buildings based on a deep convolutional neural network (CNN) algorithm. We also compare our approach to another state-of-the-art baseline model for damage mapping. The results indicated that our dataset, together with a deep learning network, enabled acceptable predictions for all the data modality scenarios, in which our approach consistently outperformed the baseline model. We also found that the results from a cross-modal mapping were comparable to the results obtained from a fusion sensor and optical mode analysis.
\end{abstract}

\begin{keyword}
Multisensor remote sensing \sep disaster damage mapping \sep deep convolutional neural network
\end{keyword}

\end{frontmatter}

%\linenumbers

\section{Introduction}
\label{S:1}
%=======================================================================

\begin{figure*}[h]
\centering\includegraphics[width=0.9\linewidth]{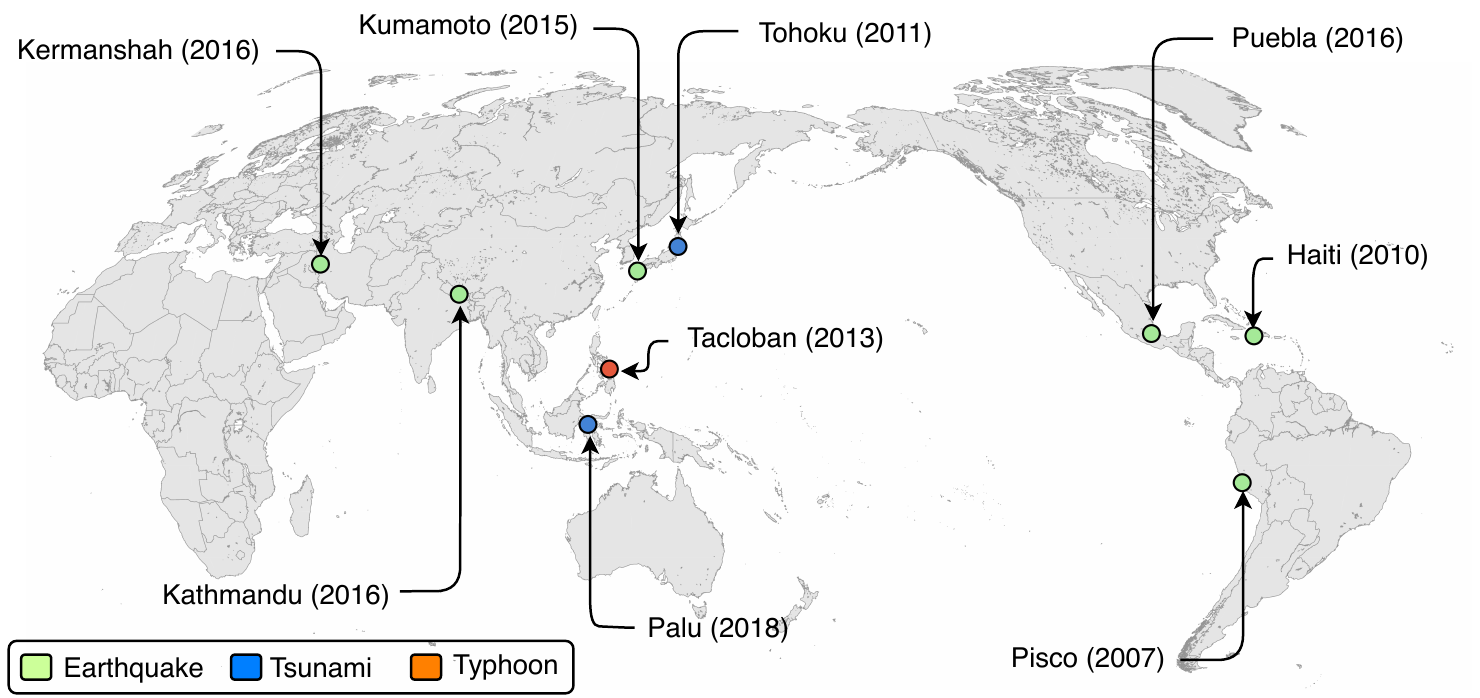}
\caption{Location of the catastrophic earthquake, tsunami, and typhoon events used to construct the multitemporal and multisensor remote sensing dataset for building damage mapping.}
\label{events_plot}
\end{figure*}

Geophysical disasters such as earthquakes and tsunamis are rare events that can devastate large urban environments, causing enormous human and economic losses. Between only 1998$-$2017, these two types of events were responsible for approximately 750,000 deaths worldwide~\citep{unreport}. Detailed information about the extent and level of structural damage is, therefore, essential to first responders for adequately conducting rescue and relief actions. In this context, earth observation (EO) technologies, such as optical imaging and synthetic aperture radar (SAR), can provide complementary information on the damage condition after a large-scale disaster~\citep{Yanbing2017, Ge2020}.

In this paper, we construct a novel global multimodal and multitemporal remote sensing dataset from notable earthquakes, tsunamis, and typhoon disasters, together with the corresponding reference data of damaged buildings. The dataset was collected from different optical sensors as well as diverse microwave ranges of the SAR data. Considering that earthquakes and tsunami events are rare occurrence events and that affected areas often become isolated because of access difficulties, collecting reliable ground-truth information is highly expensive. Thus, the advantage of the proposed framework is introducing a unique EO building damage dataset (BDD) involving mapping. By using this dataset, it becomes possible to analyze diverse scenarios of data availability, such as single-mode, cross-modal, and mode fusion data scenarios, for building damage recognition. Furthermore, we introduce a damage mapping framework for the classification of building damage from space using modern deep learning algorithms. The main contribution of this work is threefold:

\begin{itemize}
\item We construct a unique global multitemporal and multimodal EO dataset together with labeled building footprints from large-scale earthquake and tsunami events worldwide.
\item We propose a damage mapping framework that integrates remote sensing and deep learning to classify the level of building damage considering several scenarios of data availability.
\item We conduct extensive experiments and evaluate the performance of the proposed framework with other state-of-the-art deep learning approaches used for damage recognition.
\end{itemize}

\subsection{Related work}
\label{S:1.1}
% --------------------------
Building damage mapping using remote sensing datasets has been extensively studied. We can broadly divide mapping frameworks based on the EO data used. The first generation of moderate-resolution optical sensors (e.g., \href{https://www.nasa.gov/mission_pages/landsat/main/index.html}{Landsat} and the \href{https://asterweb.jpl.nasa.gov/}{Advanced Spaceborne Thermal Emission and Reflection Radiometer (ASTER)}) allowed a general interpretation of the structural damage in affected areas~\citep{Yusuf2001, Yamazaki2007}. The follow-up generation of high-resolution (HR) optical sensors enabled detailed damage recognition. Using pixel- or object-based change detection techniques, it became possible to adequately classify several degrees of damage for a single building~\citep{Freire2014}. These works were successfully applied to large disasters. For instance, following the 2008 Wenchuan earthquake, \cite{Tong2012} extracted individual collapsed buildings using 3D geometric changes (building heights) observed in pre- and post-event IKONOS images.

On the other hand, very-high-resolution (VHR) optical imagery is also used to visually interpret the damage condition after disasters. Although an experienced human interpreter can provide very reliable information, these approaches are time consuming. As such, it is mainly utilized by large international organizations, such as \href{https://www.unitar.org/maps}{the United Nations Institute for Training and Research Operational Satellite Applications Programme (UNOSAT)} and \href{https://emergency.copernicus.eu/}{Copernicus Emergency Management Service (EMS)}. Furthermore, analysis using optical imaging is often hampered by weather conditions when clouds could cause occluded acquisitions. SAR, which can penetrate clouds, has gained more popularity for disaster response tasks. Similar to optical imagery, the details of the analysis are correlated to the SAR pixel resolution. The latest SAR sensors, based on high-frequency X- and C-bands, can detect specific geometric features from built-up areas~\citep{Ferro2013, 7128381}. Polarimetric SAR is also used to extract information with different SAR scattering mechanisms that can be linked to the degree of building damage. For instance, \cite{Yamaguchi2012} demonstrated that horizontal-horizontal (HH)-polarized double-bounce data provides better features for analyzing collapsed structures. A combination of optical and SAR data was also extensively studied. \cite{Brunner2010a} used pre-disaster optical images to extract geometric parameters of isolated collapsed buildings, and then damage grading was performed by comparing a simulated post-event SAR scene, the collapsed structure, and the actual HR post-disaster SAR data.

\cite{Dong2013, rs6064870, Ge2020, Koshimura2020} presented comprehensive reviews of remote sensing for disaster mapping. They concluded that although considerable progress has been made in damage recognition from space, the success of the developed techniques mainly depends on $i)$ the quality of the validation data, which is often limited, and $ii)$ an appropriated set of pre- and post-disaster images. These factors have influenced the design of previous methods, limiting their applicability to specific characteristics of datasets and affected areas. These facts make field surveys the gold standard to obtain precise information on the damage condition. Nevertheless, the large EO dataset acquired from previous events present a valuable resource for developing advanced mapping frameworks to assess future disasters.

Recently, deep learning algorithms, such as deep convolutional neural networks (CNNs), have made significant progress in solving computer vision problems, such as object classification and image segmentation~\citep{10.1007/s10462-018-9641-3}. Due to this success, CNN algorithms have been used for damage recognition using remote sensing datasets. Recent applications verify the potential of this technology~\citep{8126255, Ma2019, rs12121924}. However, CNN models require a large number of training images with corresponding high-quality labeled data. For this reason, research groups have recently collected large datasets for a variety of tasks, such as common object detection and recognition~\citep{Lin2014, Everingham2015}.

In the field of disaster management, the xView2 Challenge~\citep{gupta2019xbd}, in collaboration with several agencies, introduced one of the first large-scale multitemporal datasets for building damage mapping. This dataset contains VHR optical imagery acquired from several disasters, such as floods, wide fires, and earthquakes. Although weather conditions may constrain the applicability of this dataset in future disasters, the research community has provided exceptional contributions in the forms of CNN architecture design and training strategies through this competition. Recently, a multimodal optical and SAR dataset was also presented in the SpaceNet-6 Challenge~\citep{shermeyer2020}. This competition was to try to extract only building footprints in a cross-modal data scheme. In this paper, we also introduce a novel multimodal and multitemporal BDD. Here, we seek to satisfy all possible data conditions that first responders face in emergency response applications.

\begin{figure*}[h]
\centering\includegraphics[width=0.85\linewidth]{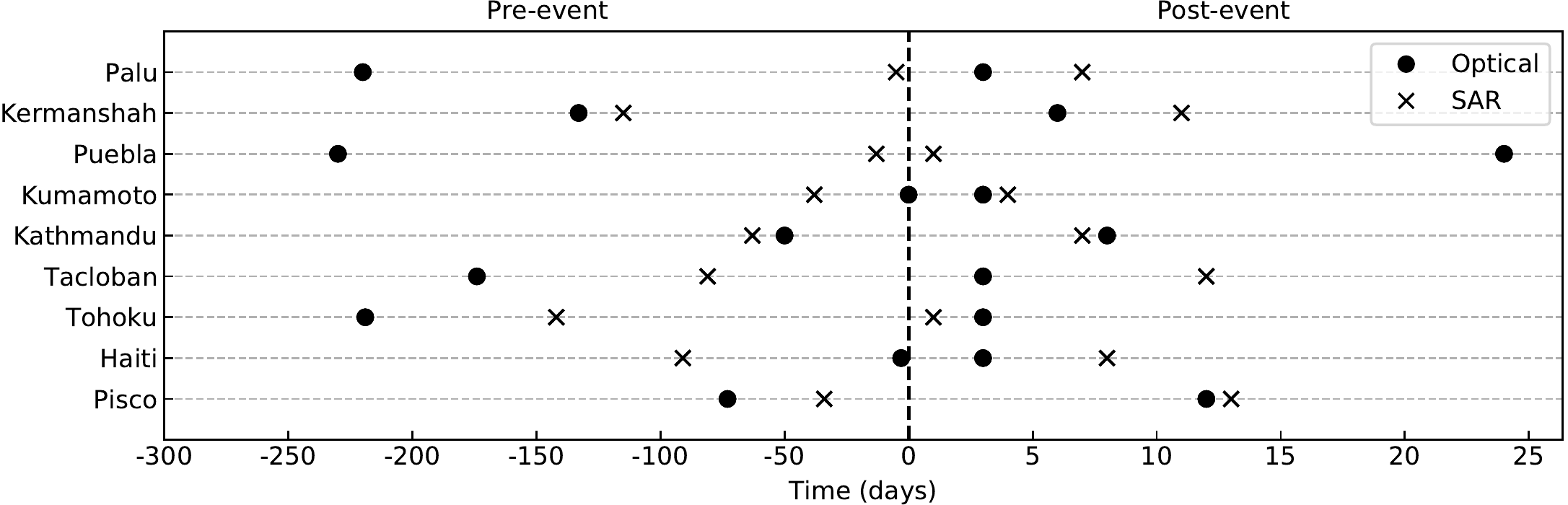}
\caption{Data acquisition times (the time difference with respect to the event origin time is in days)}
\label{events_time}
\end{figure*}

The rest of the paper is organized as follows. Section~\ref{S:2} details the remote sensing dataset and the processing and generation of the labeled building masks for segmentation of damaged buildings. Section~\ref{S:3} presents our proposed methodology, in which details of the CNN architecture and training settings are described. Sections~\ref{S:4} and~\ref{S:5} show the corresponding experimental results and the discussion, respectively. Finally, Section~\ref{S:6} provides the concluding remarks and outlook of our proposed dataset.

%=======================================================================
\section{Materials}
\label{S:2}
%=======================================================================
For a detailed classification of damaged buildings, HR or moderate-resolution remote sensing imagery is necessary. Optical imaging enables straightforward interpretation of affected areas; however, weather and daylight conditions might limit cloudless acquisitions. Recently, microwave SAR data, with nearly all-weather observation capabilities, have become an essential tool in rapid disaster response efforts. In this context, we have processed optical and SAR imagery from large-scale earthquakes, tsunamis, and typhoon disasters together with the corresponding recorded building damage. This BDD, to the best of the authors' knowledge, represents the first multimodal and multitemporal EO dataset for disaster management research.

\begin{table*}[h]
\centering
\caption{Large-scale disasters used in this study.}
\begin{tabular}{lcccm{6cm}}
\hline 
\textbf{Event} & \textbf{Disaster} & \textbf{Date} & \textbf{Country} & \textbf{Representative building damage studies} \\
\hline 
Pisco       & Earthquake & 2007-8-15  & Peru        & \cite{eeri_pisco, Taucer2009, Matsuoka_2013jdr} \\
Haiti       & Earthquake & 2010-1-12  & Haiti       & \cite{Ghosh2011, Edmund2011, Miura2016}  \\
Tohoku      & Tsunami    & 2011-3-11  & Japan       & \cite{mori2011, mori2012, Gokon2012}  \\
Haiyan      & Typhoon    & 2013-11-3  & Philippines & \cite{tajima2014, Mas2015, Roeber2015}  \\
Nepal       & Earthquake & 2015-4-15  & Nepal       & \cite{Goda2015, SHARMA201661, OKAMURA20151015} \\
Kumamoto    & Earthquake & 2016-4-14  & Japan       & \cite{Yamanaka2016, Yamada2017, Naito2020} \\
Puebla      & Earthquake & 2017-9-19  & Mexico      & \cite{ALBERTO20181073, Roeslin2018, Celebi2018} \\
Kermanshah  & Earthquake & 2017-11-12 & Iran        & \cite{Hossein2018, Vetr2018, rs10081255} \\
Palu        & Tsunami    & 2018-9-28  & Indonesia   & \cite{rs11070886, Paulik2019, Widiyanto2019} \\
\hline
\end{tabular}
\label{events_desc}
\end{table*}
\begin{table*}[h]
\centering
\caption{Available multitemporal, multimodal remote sensing dataset and the reference data of the building damage for each event.}
\begin{tabular}{lcccc}

\hline
\multirow{2}{*}{\textbf{Event}} & 
\multicolumn{2}{c}{\textbf{Remote sensing data}} &
%\multirow{2}{4em}{\textbf{Reference source}} & 
\multicolumn{2}{c}{\textbf{Building damage dataset}} \\

\cline{2-5}
            & \textbf{Optical} & \textbf{SAR} & \textbf{Source} & \textbf{Polygons} \\
\hline
Pisco       & QuickBird     & ALOS         & CISMID         &  3,164 \\
Haiti       & WorldView-2/3 & TerraSAR-X   & UNOSAT         &  2,036 \\
Tohoku      & WorldView-2/3 & TerraSAR-X   & MLIT           & 14,047 \\
Haiyan      & WorldView-2   & COSMO-SkyMed & JICA           & 21,196 \\
Nepal       & SPOT 6/7      & ALOS-2       & UNOSAT         &  1,710 \\
Kumamoto    & Pleiades      & TerraSAR-X   & GSI            & 11,469 \\
Puebla      & SPOT 6/7      & ALOS-2       & UNOSAT         &    777 \\
Kermanshah  & WorldView-2/3 & ALOS-2       & UNOSAT         &  1,052 \\
Palu        & WorldView-2/3 & COSMO-SkyMed & Copernicus EMS &  5,745 \\
\hline
\end{tabular}
\label{events_sensors}
\end{table*}

\subsection{Disaster events}
\label{S:2.1}
% -----------------------------------
For this study, we select and introduce the disasters that have a full set of multitemporal and multimodal datasets. Fig.~\ref{events_plot} shows the location of the events considered in this study. Our dataset is composed of one typhoon, six earthquakes, and two tsunami disasters. It is essential to note that only the 2011 Tohoku tsunami and the 2015 Kumamoto earthquake occurred in two distant cities in the same country. The rest of the events impacted other urban environments in diverse geographical locations. This unique characteristic provides broad information on several affected areas, considering the type of building damage and geographic conditions. Table~\ref{events_desc} lists the main characteristics of the disaster events included in this study.

\begin{figure}[h]
\centering\includegraphics[width=0.97\linewidth]{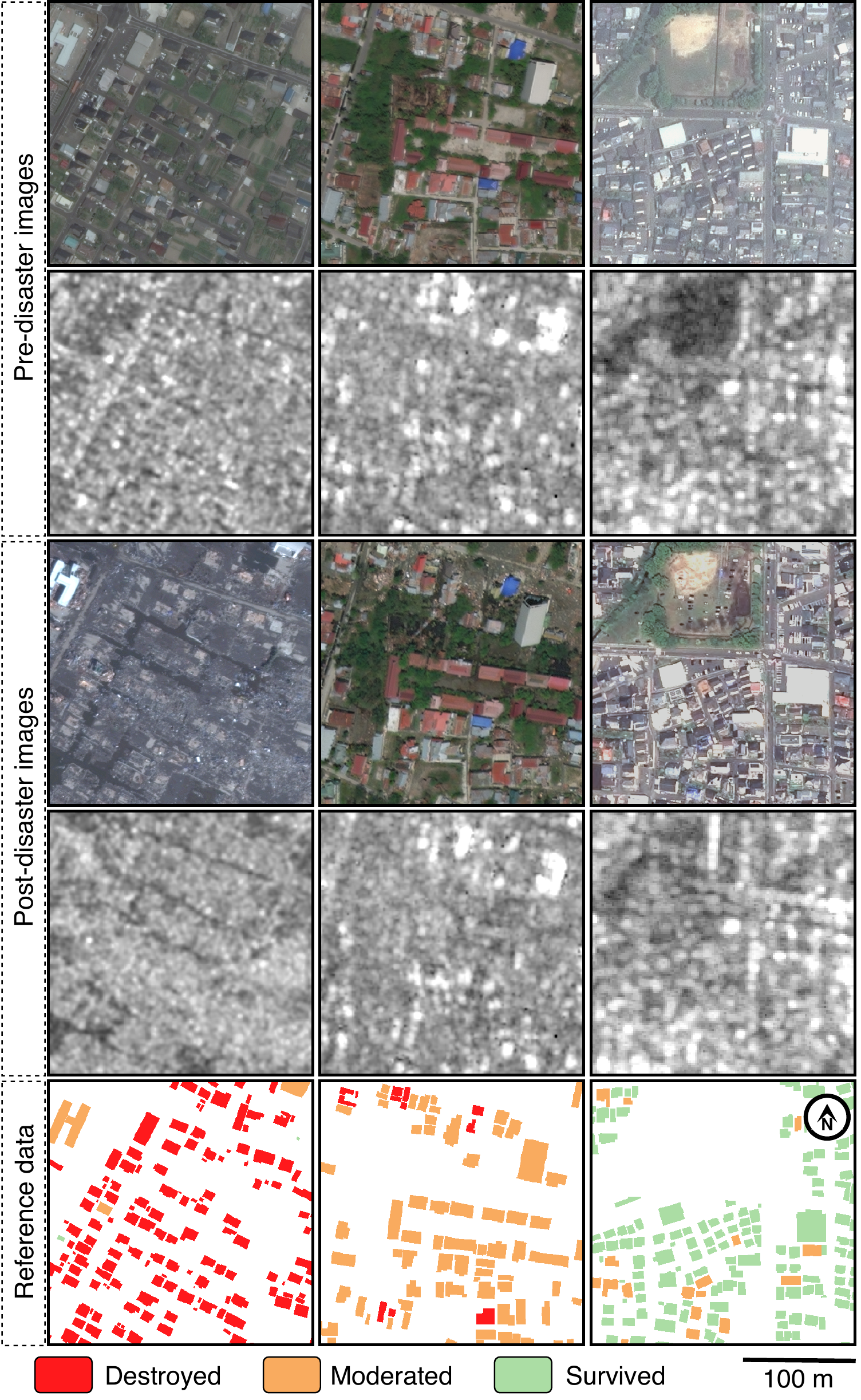}
\caption{Examples from the multitemporal and multisensor building damage dataset. The first, second and third columns show damage from earthquake, tsunami, and typhoon disasters, respectively.}
\label{eg_patches}
\end{figure}

\subsection{Optical imagery}
\label{S:2.2}
% --------------------------
Table~\ref{events_sensors} lists the optical sensors used in this study. The WorldView-2/3, QuickBird, and Pleiades sensors provide VHR images with approximately 0.5 m of ground sampling distance. On the other hand, after preprocessing (pansharpening), the pixel resolution for images from the Système Pour l'Observation de la Terre (SPOT)-6/7 sensor is approximately 1.5 m. All the images were acquired in GeoTIFF format. In this paper, we use only the spectral bands available across all events. Thus, the red, green, and blue (RGB) bands from the visible range were selected. To facilitate a change detection analysis, a set of pre- and post-event images were processed for all events. The different acquisition times (in days from the event origin time) for each disaster are depicted in Fig.~\ref{events_time}. We tried to collect images under the same season conditions. However, considering the difficulty of obtaining perfect cloud-free optical images soon before and after the event, most of the pre-event imagery was taken two to six months before the events. In the case of the post-event imagery, it was possible to process images taken within two weeks after the disaster.

Three preprocessing steps were conducted on all the multitemporal images. First, the digital number was converted to reflectance. Given the variety of sensors used and acquisition dates, several pairs of images showed a global shifting (misregistration) for several events. To address this issue, we coregister the post-event dataset using the pre-event dataset as the primary image. Finally, all the geocoded images were standardized to an 8-bit data format.

\subsection{Synthetic aperture radar}
\label{S:2.3}
% -----------------------------------
The almost all-weather acquisition capabilities of SAR sensors represent an advantage compared to optical imaging. To complement the optical dataset, we also collected a set of pre- and post-event SAR data for all events included in this study. Similar to the optical dataset, several commercial sensors provided the SAR information (Table~\ref{events_sensors}). Moreover, to take advantage of SAR data over built-up areas~\citep{Yamaguchi2012, Ferro2013}, we select the HH polarization scenes for all events. The StreetMap (SM) acquisition configuration of the TerraSAR-X platform (managed by the German Aerospace Center) and the COnstellation of small Satellites for the Mediterranean basin Observation (COSMO)-SkyMed platform (managed by the Italian Space Agency) capture HR X-band data with approximately 1.2~m and 3.3~m for the slant range and azimuth direction, respectively. On the other hand, the SM model of the Advanced Land Observing Satellite (ALOS)-2 platform captures L-band data with pixel spacings of 2.5~m and 3.15~m for the ultrafine and high-sensitivity configurations.

Several preprocessing techniques were applied to the SAR dataset. In the case of TerraSAR-X, all the SAR scenes were provided as enhanced ellipsoid corrected (EEC) products. Accordingly, to obtain the geocoded intensity images, radiometric corrections were directly applied using the local incident angle provided by the sensor. In the case of COSMO-SkyMed and ALOS-2 scenes, all the images were acquired as single-look complex (SLC) products. Thus, we conduct almost identical preprocessing steps to both sensors. The post-event SAR scenes were set as secondary images and coregistered to the pre-event scenes (primary images). In the multilooking process, we used the minimum number of looks to obtain the highest pixel resolution for each sensor. Next, to suppress SAR's speckle noise, we apply the enhanced Lee filter~\citep{803320} to the radiometrically corrected intensity images using a moving window of 3$\times$3 pixels. Finally, we use the 1-arcsecond Shuttle Radar Topography Mission (SRTM) digital elevation model (DEM) for orthorectification (terrain correction) and geocoding of all the SAR scenes.

\subsection{Generation of labels for damage categories}
\label{S:2.4}
% ----------------------------------------------------
Collecting ground-truth data of the damage condition after a large-scale disaster strikes an urban environment is costly and time consuming. However, it is an essential task, particularly if we consider that detail-labeled building units are necessary for adequately planning rescue missions. Traditionally, two main approaches are employed to gather adequate information for labeling building damage. The first method is field surveys conducted by experts that often demand a large amount of economic and human resources~\citep{Masi2017, Monfort2019}. This approach, however, can provide the most reliable information by categorizing several degrees of building damage. In the cases of the 2007 Pisco earthquake, the 2011 Tohoku tsunami, and the 2016 Kumamoto earthquake, the initial building damage classifications were constructed through field survey campaigns by the Japan-Peru Center for Earthquake Engineering Research and Disaster Mitigation (CISMID), the Ministry of Land, Infrastructure, Transport and Tourism of Japan (MLIT), and the Geospatial Information Authority of Japan (GSI), respectively~(Table~\ref{events_sensors}).

On the other hand, visually interpreting HR optical images also provides adequate information on the overall damage~\citep{10.1193/1.2101807, Koshimura2009, Gokon2012, Mas2015}. However, the details of damage interpretation are often limited because of the almost nadir-looking nature of optical sensors. For this study, we downloaded the visual damage interpretation for the 2010 Haiti earthquake, 2015 Nepal earthquake, 2017 Puebla earthquake, and 2017 Kermanshah earthquake from UNOSAT. In the case of the 2013 Haiyan typhoon, the building damage interpretation was conducted by the Japan International Cooperation Agency (JICA). Finally, the Copernicus EMS conducted a visual analysis of building damage following the 2018 Palu tsunami~(Table~\ref{events_sensors}).

\begin{table}[h]
\centering
\caption{Descriptions of the damage classes defined in this study.}
\begin{tabular}{m{4.5em}cm{9em}}
\hline
\textbf{Damage level} & \textbf{Buildings} & \textbf{Description} \\
\hline  
\cellcolor[rgb]{1.000,0.098,0.110}Destroyed & 16,542 & Completely collapsed or washed away \\
\hdashline
\cellcolor[rgb]{0.974,0.671,0.373}Moderated & 28,112 & Visible changes in and around the building \\
\hdashline
\cellcolor[rgb]{0.671,0.867,0.643}Survived  & 78,799 & The building appears undisturbed\\
\hline  
\end{tabular}
\label{damg_classes}
\end{table}

In the case of the field survey, vector files of building polygons were available for each event. However, visually interpreted data provide only the point location of the damaged building. In such cases, we used the building polygon layer of the \href{https://www.openstreetmap.org/#map=5/35.588/134.380}{OpenStreetMap} Project. Then, we overlapped the polygon and point vector layers and assigned the interpreted damage using majority voting of the points located within a building polygon~\citep{rs11070886}.

Finally, to facilitate the comparison of the different damage classification levels, we defined a three-category scale of building damage~(Table~\ref{damg_classes}). Our damage definition is based on the building's structural condition after the disaster. As such, the height degree of damage $Destroyed$ is assigned when the structure is destroyed (i.e., collapsed) or washed away. Conversely, the $Survived$ class is set when the building structure appears to be undisturbed, or there are no visible damage to the building's rooftop. Finally, the middle damage category, moderately damaged (labeled as $Moderated$), corresponds to buildings showing visible changes in their structure or surroundings. For instance, in the case of tsunami impact, debris or remaining water are visible around buildings and in scenarios of earthquake-induced damage, part of a house's roof or sidewall is damaged. Table~\ref{damg_classes} summarizes the damage classes used in this study and the number of buildings included in each class. Fig.~\ref{eg_patches} shows samples of the optical, SAR, and corresponding labeled building footprint included in the proposed BDD.

\begin{figure*}[h]
\centering\includegraphics[width=0.97\linewidth]{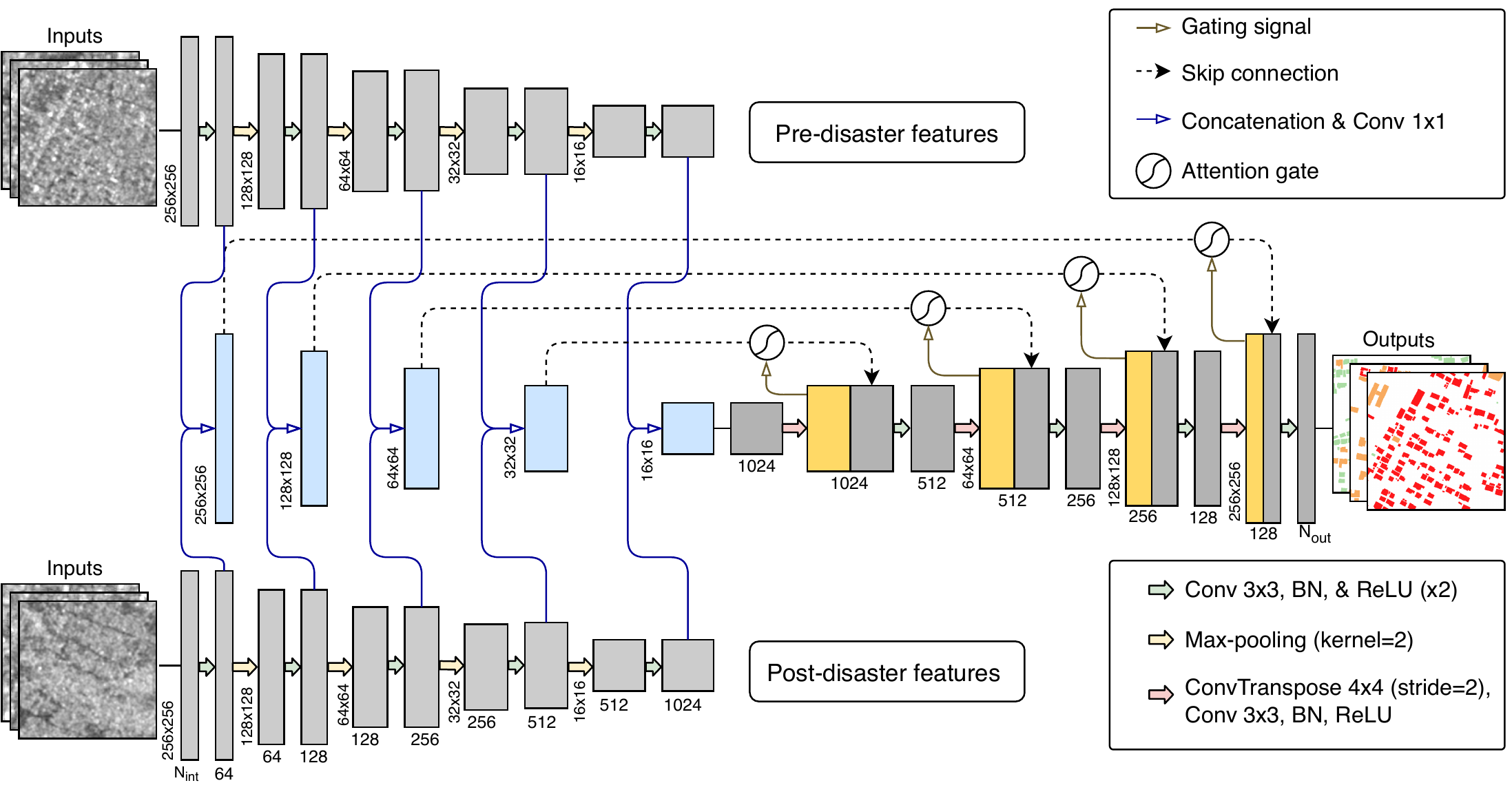}
\caption{Overview of the building damage mapping framework based on the Attention U-Net segmentation model. N$_{int}$ and N$_{out}$ denote the number of channels of the input and output images, respectively. In the case of this diagram, the input images correspond to the data mode-2 scenario (multitemporal SAR imagery).}
\label{network}
\end{figure*}

%=======================================================================
\section{Method}
\label{S:3}
%=======================================================================
We propose a framework for building damage mapping using CNNs~\citep{726791}. Given that adequate building locations are often unknown in immediate emergency response, we use a CNN model for a multiclass semantic segmentation of damaged buildings. Fig.~\ref{network} depicts the workflow of our framework. Here, the model extracts high-dimensional features from each temporal dataset separately. Then, the extracted feature vectors are used to map and grade the damaged buildings in the affected area.

\subsection{Convolutional neural network model}
\label{s:3.1}
% --------------------------------
The CNN architecture is based on a U-Net model~\citep{Ronneberger2015}. This architecture consists of an encoder-decoder design for semantic segmentation. In these types of networks, high-dimensional feature vectors are extracted from input images by the decoder using successive blocks. In this work, we modify the encoder design by adopting two encoder streams to derive features from the pre- and post-disaster datasets separately. By setting a change detection approach, the encoders share their extracted features through concatenation and 2D convolution operations (Fig.~\ref{network}).

Each encoder stream is composed of five blocks. We use a set of two 3$\times$3 2D convolutions, batch normalization, and the rectified linear unit (ReLU) activation function in each block. A 2$\times$2 max-pooling downsampling operation (kernel = 2) follows each encoder block. The output number of feature channels is doubled starting from 64 at the end of each block.

The decoder part follows a mirror design of the encoders, where the max-pooling operation is replaced by a 4$\times$4 transpose convolution to sequentially upsample the extracted feature vectors. Furthermore, to recover the original pixel resolution of the input image and to share the information learned by the encoder to the decoder~\citep{Ronneberger2015}, the U-Net model uses skip connections that concatenate the feature vectors of two corresponding encoder and decoder blocks.

As shown in Fig.~\ref{network}, in our proposed network, the encoder part shares the combined pre- and post-disaster features through the skip connections. Moreover, considering that our BDD contains diverse building structures and urban layouts, we also incorporate an additional attention gate operation, which automatically learns to focus on different target shapes by suppressing trivial regions in the input images~\cite{Oktay2018}.

To obtain the desired number of classes $N$ (Eq.~\ref{eq1}) from the last decoder block, we apply a softmax activation function to the output features vector $z$. As a result, the network outputs $N$-channel vectors with a predicted probability $p(z)$ of each class $i$ (Eq.~\ref{eq2}). Then, we compute the final categorical output $\bar{y}$ for the given input images by maximizing $p(z)$ (Eq.~\ref{eq3}).

\begin{equation}
\label{eq1}
\sigma(z)_i = \frac{e^{z_i}}{\sum^N_{n=1}e^z_n}
\end{equation}

\begin{equation}
\label{eq2}
p(z) = \left[p_1(z),\: p_2(z),...,\: p_i(z),...,\: p_N(z)\right]
\end{equation}

\begin{equation}
\label{eq3}
\bar{y} = argmax \: p(z)
\end{equation}

\subsection{Training settings}
\label{s:3.2}
% --------------------------------
U-Net-based models are known for their ability to work from relatively small training data~\citep{Ronneberger2015}. Compared with other CNN-based architectures (e.g., fully convolutional networks), U-Net is a lightweight design involving approximately $8\times10^6$ trainable parameters. However, to improve the training speed and achieve better convergence, a common technique is initializing the first layer weights using a pretrained network on a larger dataset. Thus, we adopt this strategy and fine-tune the weights of a ResNeXt~\citep{xie2016aggregated} pretrained on ImageNet.

The training process is optimized using the adaptive moment estimation (Adam) algorithm~\citep{kingma2014adam}. We use all the default parameters with an initial learning rate of $1\times10^{-4}$. The Adam algorithm adaptively computes the learning rate. However, to improve the convergence speed of hyperparameter tuning, we also use a traditional learning rate decay~\citep{7926641}. Lastly, to evaluate the performance of the network, we use the categorical cross-entropy loss (Eq.~\ref{eq4}) computed from the target labeled image $y_i$ (ground truth) and the predicted class probabilities $\bar{y}_i$ for each class. During the network training process, this loss function is gradually optimized by the Adam algorithm.

\begin{equation}
\label{eq4}
\mathcal{L_C}(y,\bar{y}) = -\sum y_i \log{\bar{y_i}}
\end{equation}

\subsection{Baseline convolutional neural network model}
\label{s:3.4}
% ------------------------------------
To evaluate our proposed BDD and CNN model, we also apply the model implemented by the winner of the \href{https://www.xview2.org/}{xView2} Challenge. This model was developed for a four-class problem using only multitemporal optical images. Here, we adapt the winning solution, with minimal modifications, for our BDD and consider a three-class semantic segmentation problem.

The winning solution from the xView2 Challenge is also based on a U-Net architecture. However, it follows a Siamese design where two separate networks are used for each pre- and post-disaster dataset. Then, the output features from the last decoder block are combined for the building damage grading task. Further details regarding the optimization algorithm and loss function settings can be found at \url{https://github.com/DIUx-xView/xView2_first_place}.

\subsection{Scenarios of data modality}
\label{s:3.5}
% ------------------------------------
Change-detection-based techniques using images taken under almost identical acquisition conditions before and after disasters have been suggested to be the most appropriate method for building damage assessment~\citep{Brett2013, 7042770}. In such methods, to facilitate an accurate change detection analysis, the post-disaster images should share similar characteristics (e.g., the SAR incident angle or season) with the pre-event images. However, as illustrated in Fig.~\ref{events_time}, it is almost unpredictable what kind of data modality will be available for an emergency response when a disaster suddenly occurs. Thus, we define five scenarios considering single-mode, cross-modal, and data fusion modes to address all possible data conditions for damage mapping. Table~\ref{datamodes} lists all the data modes for building damage classification.

\begin{table}[h]
\caption{Scenarios of data modalities.}
\centering
\begin{tabular}{ccccc} 
\hline
\multirow{2}{*}{\textbf{Mode}} & \multicolumn{2}{c}{\textbf{Pre-event}} & \multicolumn{2}{c}{\textbf{Post-event}} \\
\cline{2-5}
  & Optical      & SAR          & Optical      & SAR          \\ 
\hline
1 & $\checkmark$ & $\checkmark$ & $\checkmark$ & $\checkmark$ \\ 
\hdashline
2 & $\checkmark$ &              & $\checkmark$ &              \\ 
\hdashline
3 &              & $\checkmark$ &              & $\checkmark$ \\ 
\hdashline
4 & $\checkmark$ &              &              & $\checkmark$ \\ 
\hdashline
5 & $\checkmark$ & $\checkmark$ &              & $\checkmark$ \\ 
\hline
\label{datamodes}
\end{tabular}
\end{table}

\begin{itemize}
\item The first mode corresponds to a data fusion scenario when a set of pre- and post-event optical and SAR images are available. This scenario is under a perfect condition where optical and SAR datasets are available for emergency response.
\item The second mode is a single-mode situation based on only optical imagery. Considering the drawbacks of optical sensors (weather and lighting conditions), this scenario is also an ideal situation. Note that the xView2 Challenge was designed for this data scenario.
\item  The third mode is also a single-mode situation of SAR datasets. Recently, this scenario has been adopted for emergency response~\citep{Ge2020}.
\item The fourth mode is a cross-mode scenario using pre-event optical images and post-event SAR data. This dataset configuration is applied to the post-event SAR data that become soon accessible after the disaster or when cloud-free post-event optical images are not available.
\item The fifth mode is also a cross-mode scenario using pre-event optical and SAR images and single post-event SAR data. Here, in addition to the conditions of the fourth mode, we evaluate the strategy of using all reachable pre-disaster datasets. Thus, the scenario is an extension of the third mode.
\end{itemize}

Note that we do not include a cross-modal scenario of pre-disaster SAR data and post-disaster optical images. The continuous acquisition of visual imagery (also SAR data) by several spaceborne platforms will guarantee the accessibility of pre-disaster images for emergency response in the case of future disasters. Thus, cross-mode scenarios of pre-event SAR and post-event optical can quickly become the fourth and fifth modes.

%=======================================================================
\section{Results}
\label{S:4}
%=======================================================================
In this section, we describe the settings of our numerical experiments and report the results from the proposed scenarios of data modality (Table~\ref{datamodes}) using our proposed mapping framework and the baseline model.

\subsection{Experimental settings}
\label{S:4.1}
% --------------------------------
All the experiments follow supervised machine learning settings. Here, using a training dataset, the CNN model learns a nonlinear mapping from labeled samples to corresponding feature vectors. For semantic segmentation applications, the labeled data correspond to pixel masks (rasterized from the building polygons with assigned damage classes), and the feature vectors are the multiband EO images. Throughout training the CNN model, a separate validation dataset is used to evaluate, in an unbiased manner, the model fit on the training dataset. Finally, the generalization ability of the trained model is assessed using a test dataset, which is independent of the training and validation datasets.

In this study, we crop the processed GeoTIFF images and corresponding labeled raster masks into tiles of 250$\times$250~m$^2$ and select tiles with at least 5\% of built-up area. Accordingly, 1,147 tiles were collected from all the events. Fig.~\ref{eg_patches} shows examples of the extracted image tiles and their matching labeled masks. For testing the CNN model, we randomly split and hold out 10\% of the tiles. During the training/validation stage, the remaining 90\% of the dataset is randomly split into training and validation datasets at an 80:20 ratio. Here, to evaluate the model performance using diverse dataset splits and ensure generalizability, we perform separate experiments using three different random seed numbers for constructing the training and validation sets~\citep{ROGAN20082272}.

Similar to the xView2 solution, during training, we monitored the network accuracy using a harmonic mean of the $F_{score}$ metric computed for each damage class. This coefficient (Eq.~\ref{eq5}) is the harmonic mean of the fraction of positive predicted pixels (also known as the $precision$) and the sensitivity of actually predicting positive pixels (also known as the $recall$). Finally, all the models were trained with a batch size of 16 for a total of 150 epochs.

\begin{equation}
\label{eq5}
F_{score} = 2 \cdot \frac{precision \cdot recall}{precision + recall}
\end{equation}

\subsection{Numerical results}
\label{S:4.2}
% --------------------------------
The reported numbers are calculated on the hold-out test dataset. The damage grading results are computed as the ensemble of three networks trained in independent experiments defined by the random seed used to split the training and validation datasets.

\begin{figure*}[h!]
\centering\includegraphics[width=1\linewidth]{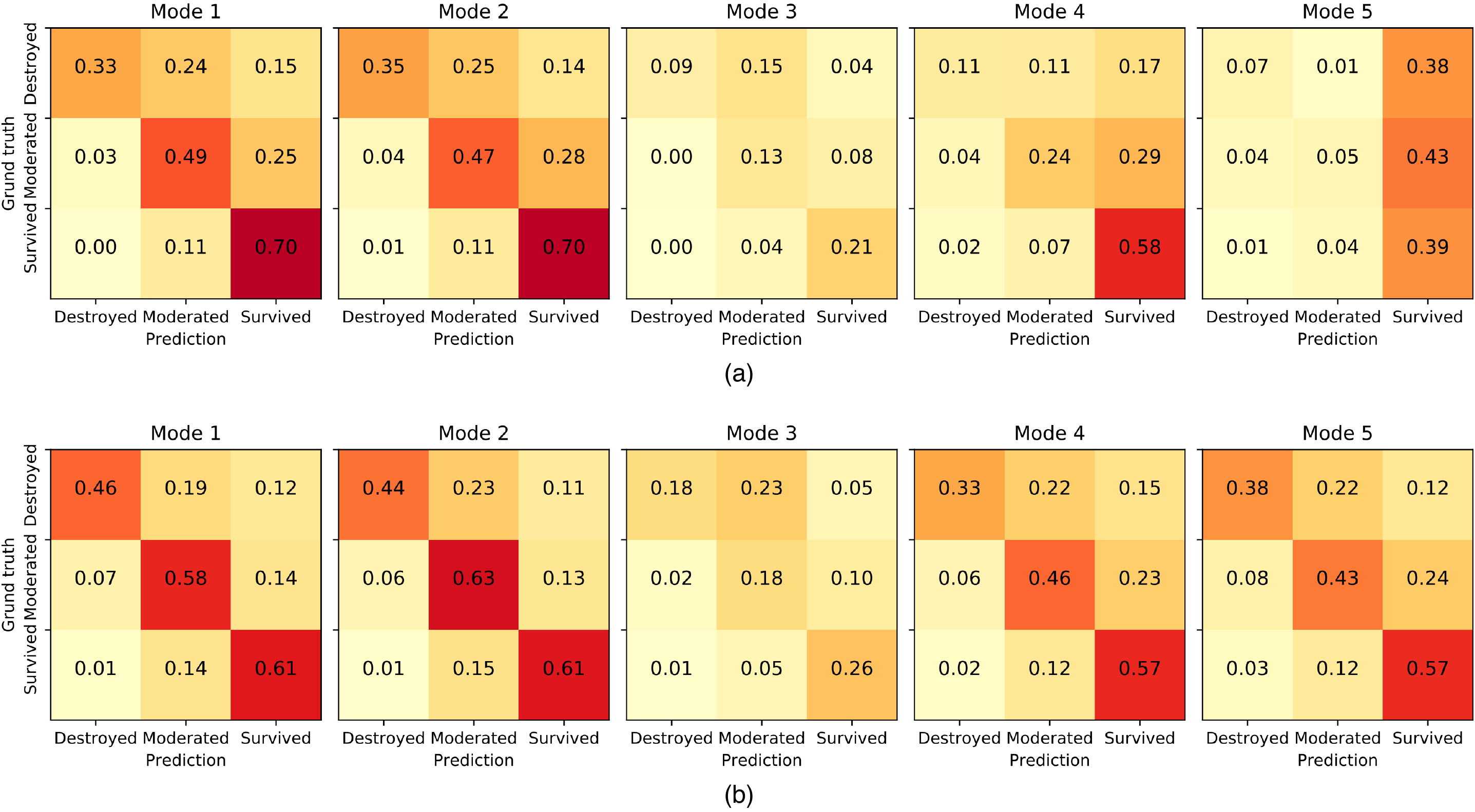}
\caption{Error matrices computed over the test tiles from each data mode. (a) shows the results of using the winning solution of the xView2 Challenge, and (b) shows the results of our proposed framework.}
\label{metrics}
\end{figure*}

Fig.~\ref{metrics}(a) shows the normalized error matrices computed on the test dataset using xView2's winning model. This figure indicates that Mode~1 (fusion of multitemporal optical and SAR imagery) and Mode~2 (only the optical images, similar to the xView-2 Challenge) achieve good performance for multiclass building damage grading. In particular, the accuracy achieved for detecting the $Survived$ class is approximately 0.70 for these data modality scenarios. However, the performance decreases to average values of 0.50 and 0.35 for the $Moderated$ and $Destroyed$ classes, respectively. Here, notable misclassifications emerge for these classes, where the CNN model confuses $Destroyed$ as the $Moderated$ class and $Moderated$ as the $Survived$ class. On the other hand, the results for Mode~3 (only the multitemporal SAR dataset) indicate a very low performance of the winning solution. In this case, the network fails to classify all three damage categories accurately. In the case of Mode~4 (cross-modal mapping using pre-disaster optical images and post-disaster SAR data), the results show that only the $Survived$ class achieves a moderate performance, with an accuracy of approximately 0.58. Finally, the fifth data mode also shows poor performance. In this case, only the $Survived$ class is slightly identified; however, the network also confuses the other damage classes with the $Survived$ class, providing an overall incorrect result for building damage mapping.

\begin{table*}[h]
\caption{Accuracy assessment ($F_{score}$) calculated on the hold-out test dataset.}
\centering
\begin{tabular}{cccccc} 
\hline
\multirow{2}{*}{\textbf{Mode}} & \multirow{2}{*}{\textbf{Method}} & \multicolumn{3}{c}{\textbf{Classes}}   & \multirow{2}{*}{\textbf{Mean}} \\
\cline{3-5}
               &           & \textbf{Destroyed} & \textbf{Moderated} & \textbf{Survived} &                    \\ 
\hline
\multirow{3}{*}{1} & xView2    & 0.3912 $\pm$ 0.0016 &  0.4396 $\pm$ 0.0079 &  \textbf{0.6197 $\pm$ 0.0048} &  0.4835 $\pm$ 0.0048 \\
& This study & \textbf{0.4173 $\pm$ 0.0053} &  \textbf{0.4567 $\pm$ 0.0092} &  0.6143 $\pm$ 0.0092 &  \textbf{0.4961 $\pm$ 0.0295} \\
\hline
\multirow{3}{*}{2} & xView2    & 0.3795 $\pm$ 0.0114 &  0.4229 $\pm$ 0.0126 &  0.6173 $\pm$ 0.0067 &  0.4732 $\pm$ 0.0102 \\
& This study & \textbf{0.4137 $\pm$ 0.0165} &  \textbf{0.4714 $\pm$ 0.0026} &  \textbf{0.6178 $\pm$ 0.0026} &  \textbf{0.5010 $\pm$ 0.0300} \\
\hline
\multirow{3}{*}{3} & xView2    & 0.1403 $\pm$ 0.0247 &  0.1768 $\pm$ 0.0198 &  0.2799 $\pm$ 0.0351 &  0.1990 $\pm$ 0.0265 \\
& This study & \textbf{0.2044 $\pm$ 0.0145} &  \textbf{0.2242 $\pm$ 0.0271} &  \textbf{0.3114 $\pm$ 0.0271} &  \textbf{0.2467 $\pm$ 0.0156} \\
\hline
\multirow{3}{*}{4} & xView2    & 0.1439 $\pm$ 0.0274 &  0.2941 $\pm$ 0.0714 &  0.5584 $\pm$ 0.0074 &  0.3321 $\pm$ 0.0354 \\
& This study & \textbf{0.3215 $\pm$ 0.0293} &  \textbf{0.3936 $\pm$ 0.0101} &  \textbf{0.5775 $\pm$ 0.0101} &  \textbf{0.4309 $\pm$ 0.0371} \\
\hline
\multirow{3}{*}{5} & xView2    & 0.1013 $\pm$ 0.0894 &  0.0845 $\pm$ 0.0344 &  0.4242 $\pm$ 0.0344 &  0.2033 $\pm$ 0.0392 \\
& This study & \textbf{0.3333 $\pm$ 0.0043} &  \textbf{0.3774 $\pm$ 0.0140} &  \textbf{0.5692 $\pm$ 0.0186} &  \textbf{0.4266 $\pm$ 0.0348} \\
\hline

\label{scores}
\end{tabular}
\end{table*}

The performance of our mapping framework for all data modes is summarized by normalized error matrices (Fig.~\ref{metrics}(b)). Similar to the baseline results (xView2 model), these results also indicate that Mode~1 and Mode~2 achieve the highest performance. Here, although the accuracy values for the $Survived$ class are slightly lower than those of the xView2 implementation, the misclassification in the three damage classes is considerably reduced. Notably, the results of using multitemporal optical datasets produce an accuracy of approximately 0.63 for classifying the intermediate damage level. This value represents the maximum accuracy among all the classes. However, false classification also occurs in this data mode where the network confuses the $Destroyed$ class as the $Moderated$ class. The latter behavior is slightly reduced when the optical and SAR datasets are fused (Mode~1), where the accuracy for the $Destroyed$ class is marginally higher than that for Mode~2.

On the other hand, the network trained on Mode~3, similar to the xView2 solution, cannot correctly distinguish all the damage classes. However, the overall accuracy values of all the damage levels is slightly higher than that of the baseline. The results from Mode~4 indicate that, in general, our approach outperforms the winning solution from the xView2 Challenge. In this mode, our network produced a more balanced classification distribution for all classes, which is comparable to the results obtained in Mode~2 by the baseline. Finally, the results of Mode~5 also indicate the superior performance of our framework. Here, the accuracies of the $Survived$ and $Moderated$ classes are comparable to those obtained in Mode~4. In this last mode, however, our network shows slightly better accuracy in predicting the $Destroyed$ class.

Table~\ref{scores} shows the overall quantitative evaluation ($F_{score}$ values and corresponding standard deviations) computed on the hold-out test dataset by two previous approaches for building damage mapping. It shows that for the two models, the xView2 model and our approach, are relatively stable across all experiments. Here, we can see that our mapping framework achieves superior results for almost all the data modality scenarios in comparison to the winning solution from the xView2 Challenge. The proposed framework, which simultaneously extracts and classifies building damage, gives the highest average scores when optical datasets are involved. The same efficiency is also observed in a cross-modal dataset (Mode~4). In the case of the SAR dataset (Mode~3), the scores demonstrate that the baseline and our model cannot produce satisfactory results. However, the results of Mode~5 (considered an extension of Mode~3) show a considerable improvement when a pre-disaster optical dataset is used for the input images.

%=======================================================================
\section{Discussion}
\label{S:5}
%=======================================================================

\begin{figure*}[h!]
\centering\includegraphics[width=0.95\linewidth]{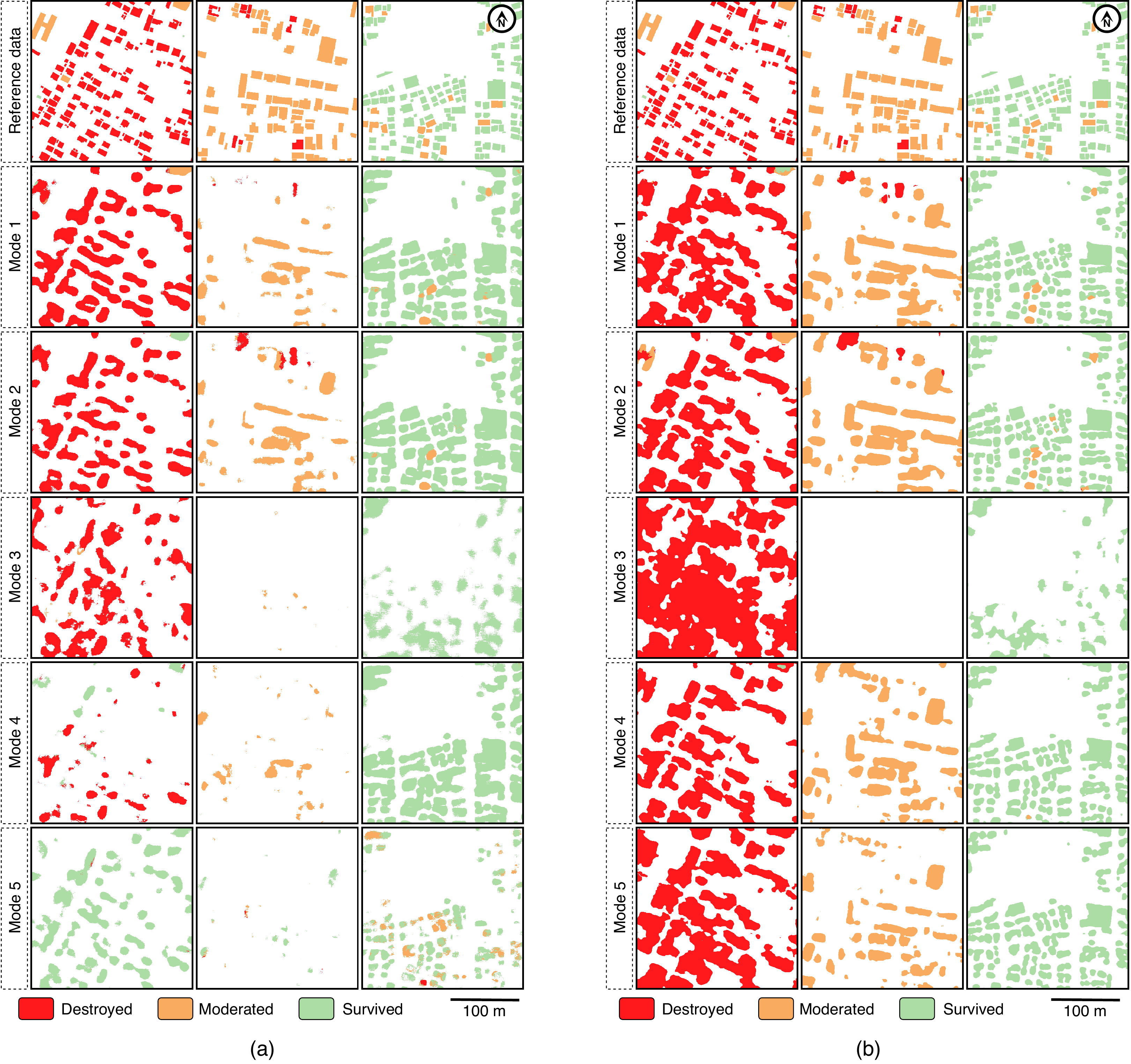}
\caption{Visual comparison of the damage classification results. (a) shows the results using the winning solution from the xView2 Challenge, and (b) shows the results of our proposed framework.}
\label{results}
\end{figure*}

In this study, we created a one-of-a-kind multitemporal and multisensor BDD from three different types of large-scale disasters, namely, earthquakes, tsunamis, and typhoons. We considered worldwide locations providing our BDD with high heterogeneity in terms of building characteristics as well as landscape configurations. Here, the reference building damage masks were predominantly constructed from visual interpretation analysis. Hence, the results obtained in this study represent a relative approximation to the human visual interpretation ability. This is particularly the case for tsunami and typhoon events where damage to sidewalls, affected by water waves or wind speeds, is typically difficult to observe using optical imagery. This latter condition highlights the importance of including SAR data, which are characterized by their side-looking observation nature. Second, given that the percentage of earthquake events is approximately 60\%, our results may also be biased, achieving higher performances for this kind of disaster. Nevertheless, we tried to reduce this bias by randomly splitting the training, validation, and testing datasets for our experiments.

Sudden disasters and weather conditions can limit our options for using an optimal set of remote sensing images for emergency response. For such circumstances, our BDD enabled extensive analysis by considering several scenarios of data availability, such as single-mode, cross-mode, and fusion-mode optical and SAR data. We presented a framework for building damage mapping based on a modern Attention U-Net architecture, showing that our approach satisfactorily classified building damage into three levels. The obtained results also indicated that our framework outperformed the baseline model (Fig.~\ref{results}(a)) based on the winning solution from the xView2 Challenge.

Our proposed framework trained a shared-encoder Attention U-Net to extract and classify building damage simultaneously. The results of this approach are shown in Fig.~\ref{results}(b). As expected, this approach achieves great results when optical imagery is included in the network's input~\citep{SHI2020184, 9082125}. Interestingly, Mode~1 and Mode~2 show almost identical performances (Figs.~\ref{results}(a)-(b) and Table~\ref{scores}). Considering that a data fusion mode of optical and SAR is used in Mode~1, these results suggest that the features derived from SAR give less information for the classification task compared with the optical-derived features. This fact is also correlated with the number of channels used in each data mode: 3 channels (RGB) and 1 channel (grayscale) for the optical and SAR data, respectively. Furthermore, according to the segmentation results of our model, Mode~1 and Mode~2 overestimate the building footprint size in the $Destroyed$ class. This effect could be reduced by incorporating an additional loss function (e.g., the Dice coefficient)~\citep{Wei2020}.

On the other hand, our framework and the baseline model do not work well using only SAR images (Mode~3). Primarily, worse predictions are obtained for the $Moderated$ and $Survived$ classes where both networks failed to delineate the building footprints accurately. The ground sampling distance of the SAR dataset ranges from 5 m to 10 m, and the predominant building size ranges from 100 m$^2$ to 200 m$^2$. These dimensions indicate that a higher resolution of SAR data is desirable for multiclass semantic segmentation damage mapping~\citep{Shahzad2019}. In the case of the $Destroyed$ class, although both networks do not fully represent the shape of the building footprints, detected pixel patterns indicated the location of destroyed buildings. This fact suggests that SAR images of our BDD could be used to reliably classify severe building damage with a different mapping scheme, for instance, using a tile-based mapping scheme.

In cases of cross-modal damage mapping (Mode~4), our framework notably outperforms the baseline results. In addition, these results also show that the CNN model, trained on our BDD, successfully executed a change detection analysis between the pre-disaster features learned from the optical dataset (e.g., building locations) and post-disaster features derived from the SAR dataset. Here, we want to emphasize that building damage classification in Mode~4 could easily become the best option for rapid emergency response~\citep{Brunner2010a, Geib2015}.

In the case of Mode~5, considered an extension of Mode~3 by using optical imagery acquired before the disaster to the pre-disaster SAR input data, the segmentation results show a remarkable performance boost (approximately 50\%) compared to Mode~3, showing almost similar results as those achieved with Mode~2 and Mode~4. Here, the pre-disaster optical data provide relevant features to teach the network to recognize $Moderated$ and $Survived$, which were not possible using only SAR datasets. Furthermore, although our results based on the SAR dataset range from 0.42-0.43, these numbers are consistent with the outcomes of the recent \href{https://spacenet.ai/sn6-challenge/}{SpaceNet Challenge 6} (the cross-modal building footprint extraction task).

%=======================================================================
\section{Conclusions}
\label{S:6}
%=======================================================================
In this study, we created a novel BDD considering three levels of damage and multiple multisensor satellite images (optical imaging and SAR data), which is notably applicable for emergency response in the case of future disasters. We presented a damage mapping framework based on an Attention U-Net architecture. This network was extensively trained, considering different and realistic scenarios of the availability of emergency disaster response data (single-mode, cross-modal, and fusion of optical and SAR datasets). In addition, we compared our results to a baseline model using a modified version of the winning solution from the xView2 Challenge. We demonstrated that our mapping framework consistently outperformed the baseline model in all data modality scenarios. We found that our network trained with optical images can accurately extract and classify building damage without any additional input (building masks). Furthermore, it was also shown that acceptable classification results could be obtained by integrating pre-disaster optical images and post-disaster SAR data.

For future research directions, we will extend the current version of our dataset by including other large-scale disasters from around the world as well as remote sensing data with several spatial resolutions. Furthermore, we plan to analyze different learning scenarios considering a disaster-wise split of the training and testing datasets. The findings of this study serve as an initial phase for developing a fully operational all-weather building damage mapping system.

% \begin{equation}
% \label{eq:emc}
% e = mc^2
% \end{equation}



%% References


\bibliographystyle{model5-names}
\bibliography{references.bib}






% \end{thebibliography}

\end{document}